%% file: kb.tex
\documentstyle[epsf]{kbsjrnl} 

\input{kmathdefs}

\begin{document}
\begin{article}

     
   \begin{kluwerdates}
     Received Date
     Accepted Date
     Final Manuscript Date
   \end{kluwerdates}
     
\journame{}
\volnumber{}
\issuenumber{}
\issuemonth{}
\volyear{}
     
\received{May 1, 1991}\revised{}
     
\authorrunninghead{}
\titlerunninghead{}
     
\setcounter{page}{159} 

\title{The Canonical Distortion Measure for Vector Quantization and
	Function Approximation\thanks{This work was supported in part
	by EPSRC grant \#K70366 and \#K70373}}
     
\author{Jonathan Baxter}
\email{jon@syseng.anu.edu.au} 
     
\affil{Department of Systems Engineering \\ Australian National University \\
Canberra 0200 \\ Australia}
     
\editor{}
     
\abstract{ To measure the quality of a set of vector quantization
points a means of measuring the distance between a random point and
its quantization is required. Common metrics such as the {\em Hamming}
and {\em Euclidean} metrics, while mathematically simple, are
inappropriate for comparing natural signals such as speech or
images. In this paper it is shown how an {\em environment} of
functions on an input space $X$ induces a {\em canonical distortion
measure} (CDM) on X.  The depiction ``canonical'' is justified because
it is shown that optimizing the reconstruction error of X with respect
to the CDM gives rise to optimal piecewise constant approximations of
the functions in the environment. The CDM is calculated in closed form
for several different function classes.  An algorithm for training
neural networks to implement the CDM is presented along with some
encouraging experimental results.}

\input{quant}

     
\bibliographystyle{plain}
\bibliography{bib}
     
\end{article}
\end{document}

%% file: kmathdefs.tex
\newcommand{\bb}{}
\newcommand{\eqref}[1]{(\ref{#1})}
\renewcommand{\Box}{\qed}

\renewcommand{\glossary}[2]{}
\def\addcontentsline#1#2#3{} 

\newtheorem{thm}{Theorem}

\newtheorem{lem}[thm]{Lemma}

\newcommand{\ie}{{\em i.e.\ }}

\newcommand{\iid}{{i.i.d.\ }}

\newcommand{\eg}{{\em e.g.\ }}

\newcommand{\ptt}{p(z|\thetat)}

\newcommand{\de}{=}

\newcommand{\Ehat}{{\hat E}}

\newcommand{\fhat}{{\hat f}}

\newcommand{\rhat}{{\hat \rho}}

\newcommand{\be}{\begin{equation*}}

\newcommand{\ep}{{\varepsilon}}

\newcommand{\M}{{\cal M}}
\newcommand{\F}{{\cal F}}

\renewcommand{\(}{\left(}
\renewcommand{\)}{\right)}
\renewcommand{\[}{\left[}
\renewcommand{\]}{\right]}
\newcommand{\text}[1]{\mbox{\rm #1}}

\renewcommand{\hbar}{{\overline h}}

\newcommand{\thetat}{{\tilde{\theta}}}

\newcommand{\xv}{{\vec x}}

\newcommand{\Xv}{{\vec X}}

\newcommand{\R}{{\bb R}}

\newcommand{\argmin}{{\mbox{argmin}}}

%% file: quant.tex
\section{Introduction} 

Consider the problems ``What are appropriate distortion measures for
images of handwritten characters, or images of faces, or
representations of speech signals?'' Simple measures such as squared
Euclidean distance, while widely used in vector quantisation
applications\cite{Graytext}, do not correlate well with our own
subjective notion of distance in these problems.  For example, an
image of an ``A'' and a slightly larger image of an ``A'' look
subjectively very similar to a human observer, although their squared
Euclidean separation (measured on a pixel by pixel basis) is very
large. The same can be said for two images of my face viewed from
slightly different angles.  Finding distortion measures that more
accurately correlate with our subjective experience is of great
practical utility in vector-quantisation and machine learning.
Quantisation with a poor distortion measure will cause the encoder to
make inefficient use of the available codebook vectors.  Also,
learning using nearest neighbour techniques or by generating a
piecewise constant approximation to the target function will be more
effective if an appropriate measure of the distortion between input
vectors is available.

From a purely philosophical perspective there is no {\em a priori}
natural distortion measure for a particular space $X$ of images or
signals. To generate a distortion measure some extra structure must be
added to the problem.  In this paper it is argued that the required
extra structure is given in the form of an {\em environment} of
functions $\F$ on $X$.  For example, consider each possible face
classifier as a function on the space of images $X$. The classifier
for ``jon'', $f_{\text{jon}}$, behaves as follows: $f_{\text{jon}}(x)=1$
if $x$ is an image of my face and $f_{\text{jon}}(x) = 0$ if $x$ is
not an image of my face. Note that $f_{\text{jon}}$ is {\em constant}
across images of the same face: it gives constant value ``1'' to
images of my face and constant value ``0'' to images of anyone else's
face. Similarly, there exists classifiers in the environment that
correspond to ``mary'' ($f_{\text{mary}}$), ``joe'' ($f_{\text{joe}}$)
and so on. All these classifiers possess the same property: they are
constant across images of similar looking faces. Thus, information
about the appropriate distortion measure to use for faces is somehow
stored in the environment of face classifiers. Similar considerations
suggest that it is the class of character classifiers that generate
the natural distortion measure for characters, it is the class of
spoken words that generate the natural distortion measure for speech
signals, it is the class of smooth functions that generate the natural
distortion measure for regression problems, and so on. A more formal
justification for this assertion will be given in section \ref{metric}
where an explicit formula is presented for the distortion measure
generated by a function class.  Such a distortion measure is termed
the {\em Canonical Distortion Measure} or CDM for the function
class. Loosely speaking, the canonical distortion between two inputs
$x$ and $x'$ is defined to be the difference between $f(x)$ and
$f(x')$, averaged over all functions $f$ in the environment.

In section \ref{proof} an optimality property of the CDM is proved,
namely that it generates optimal quantization points and Voronoi
regions for generating piecewise constant approximations of the
functions in the environment. In section \ref{examples} the CDM is
explicitly calculated for several simple environments. In particular
it is shown that the squared Euclidean distance function is the CDM
for a {\em linear environment}. This leads to the interesting
observation that the squared Euclidean distance is optimal for
approximating linear classes (and in fact {\em only} optimal for
approximating linear classes).

In section \ref{qex} the optimal quantization points for a quadratic
environment are calculated, and the relationship between the spacing
of the points and the behaviour of the functions in the environment is
discussed. 

The relationship between the CDM and other approaches to learning
distance metrics is discussed in section \ref{relation} (see also
section \ref{rel} below), where we will
see that the CDM provides a unifying conceptual framework for many
seemingly disparate threads of research. 

In section \ref{learn} it is shown how the CDM may be {\em learnt} by
sampling both from the input space {\em and} the environment, and an
(albeit toy) experiment is reported in which the CDM is learnt for a
``robot arm'' environment. The resulting CDM is then used to
facilitate the learning of piecewise constant approximations to the
functions in the environment. The same functions are also learnt
without the help of the CDM and the results are compared. At least on
this toy problem, learning with the CDM gives far better
generalisation from small training sets than learning without the
CDM. 

\subsection{Related work}
\label{rel}
Other authors have investigated the possibility for using specially
tailored distance functions in both machine learning and vector
quantization contexts. The authors of \cite{Simard} used a measure of
distance that takes into account invariance with respect to affine
transformations and thickness transformations of handwritten
characters. They achieved a notable improvement in performance using
this measure in a nearest neighbour classifier.  Independently but
simultaneously with some of the work in a previous incarnation of this
paper \cite{thesis}, the authors of \cite{Thrun} proposed an
``invariance measure'' on images that has a close relationship to the
CDM defined here, under certain restrictions on the functions in the
environment. They also presented techniques for using the invariance
measure to facilitate the learning of novel tasks. In work
that is close to the spirit of the present paper, the authors of
\cite{Gray} considered vector quantization in a Bayes classifier
environment.  They modified the usual squared Euclidean reconstruction
error (\eqref{recerr} in section \ref{vq} below, with $d =
\|\cdot\|^2$) by adding a Bayes risk term, and presented results
showing improved combined quantization and classification performance
in several real-world problems.  The relationship between all these
approaches and the CDM is discussed further in section \ref{relation}.

As an historical note, the idea of using a set of functions to
generate a pseudo-metric was used at least twenty years ago as a
technical tool in the theory of topology \cite{Negropontis}, although
that application has little to do with the present paper.

\section{The distortion measure and vector quantization}
\subsection{Vector quantization}
\label{vq}
As any real digital communication channel has only finite capacity,
transmitting continuous data (e.g speech signals or images) requires
first that such data be transformed into a discrete representation.
Typically, given a probability space $(X,\Sigma,P)$ ($\Sigma$ is a
$\sigma$-algebra of subsets of $X$ and $P$ is the probability
distribution on $X$), one chooses a {\em quantization} or codebook
$\{x_1,\dots,x_k\}\subset X$, and instead of transmitting $x\in X$,
the index of the ``nearest'' quantization point
\begin{equation}
\label{qd}
q_d(x) := \argmin_{x_i} d(x_i,x)
\end{equation}
is transmitted, where $d$ is a distortion measure (not necessarily a
metric) on $X$.  The quantization points $\xv = \{x_1,\dots,x_k\}$ are
chosen so that the expected distortion between $x$ and its
quantization $q_d(x)$ is minimal, i.e $\xv$ are chosen to minimize the
{\em reconstruction error}
\begin{equation}
\label{recerr}
E_d(\xv) \de \int_X d(x,q_d(x))\, dP(x).
\end{equation}
A common approach to minimizing \eqref{recerr} is Lloyd's
algorithm which iteratively improves a set of quantization points
based on a centroidal update (see \cite{lloyd,CT,Graytext}).

Some examples of distortion measures are the {\em Hamming metric},
$$
d(x,x') = 1 - \delta(x,x'),
$$
where $\delta$ is the Kronecker delta function, and the {\em squared
Euclidean distortion measure} for vector-valued $X$,
$$
d(x,x') = \|x-x'\|^2.
$$
The use of these distortion measures has more to do with their
convenient mathematical properties than their applicability to any
particular problem domain. For example, suppose $X$ is a space of
images and it is images of characters that are being transmitted over
the channel. An image of the character ``A'' and another translated
image of the same character would be considered ``close'' in this
context, although the squared Euclidean distance between the two
images would be large, quite likely larger than the distance between
an image of ``A'' and an image of ``B'' in the same location. Thus the
squared Euclidean distortion measure does not capture well the idea of
two images being ``close''. Another example is that of speech
coding---there is a large squared Euclidean distance between a speech
signal and a small translation of it in time, although both sound very
similar to a human observer. A vector quantizer constructed using an
inappropriate distortion measure will require a far larger set of
quantization points to achieve satisfactory performance in
environments where we are primarily interested in the {\em subjective}
quality of the reconstruction. And in almost all applications of
vector quantization it is the subjective quality of the reconstruction
that is the most important: eventually the quality of the
reconstructed speech signal or image is measured by how close it
appears to be to the original for some observer.

In the next section we will see how the problem of choosing an
appropriate distortion measure may be solved, at least {\em in
principle}, by using the idea of the {\em environment} of a
quantization process.

\subsection{The Environment}
\label{metric}
What makes the translated image of an ``A'' close to the original,
while an untranslated image of a ``B'' quite distant? And what makes
two speech signals nearly identical even though they are miles apart
from a Euclidean perspective? As discussed in the introduction, it is
because there is an {\em environment} of functions (the character
classifiers) that vary little across images of the same
character, and similarly there is an environment of ``speech
classifiers'' that vary little across utterances of the same
sentences or words.

Note that the set of {\em possible} character classifiers
is much larger than the set of particular classifiers for
the English alphabet (English language).  The particular form of the
letters we use is arbitrary (as evidenced by the existence of vastly
different alphabets such as Hebrew or Kanji), the only thing that is
required of a character is that different examples of it are
recognisably the same object (to us). Thus the number of different
character classifiers is potentially astronomical. A similar
conclusion holds for speech: the set of spoken word classifiers is
nearly infinite if one includes all possible words from all possible
languages. 

\subsection{Formal definition of the Canonical Distortion Measure (CDM)}
\label{formal}
To formalize the idea that it is an environment of functions that
determines the distortion measure on the input space, define the {\em
environment} of any probability space $(X,\Sigma,P)$ to be a pair
$(\F,Q)$ where $\F$ is a set of functions mapping $X$ into a space
$(Y,\sigma)$, where $\sigma\colon Y\times Y\to \R$ ($\sigma$ may be a
metric), and $Q$ is a probability measure on $\F$\footnote{For $Q$ to
be well defined there needs to be a $\sigma$-algebra on $\F$. We leave
that algebra unspecified in what follows, and simply assume that an
appropriate one exists.}. An environment so defined induces the
following natural distortion measure on $X$:
\begin{equation}
\label{rho}
\rho(x,x') \de \int_\F \sigma(f(x),f(x'))\, dQ(f),
\end{equation}
for all $x,x'\in X$. In words, $\rho(x,x')$ is the average distance
between $f(x)$ and $f(x')$ where $f$ is a function chosen at random
from the environment.

Note that if $\sigma$ is a metric on $Y$ then $\rho$ is a
pseudo-metric on $X$ (recall that a pseudo-metric is the same as a
metric except that $\rho(x,x') = 0$ does not necessarily imply
$x=x'$).  From now on $\rho$ will be referred to as the {\em Canonical
Distortion Measure} or {\em CDM} for the environment $(\F,Q)$.

In relation to the character transmission problem, $\F$ would consist
of all character like classifiers, $Y$ would be the set $[0,1]$ and we
could take $\sigma(y,y') = (y-y')^2$ (here we are assuming that the
functions $f \in \F$ are {\em probabilistic classifiers}, that is
$f(x)$ represents the probability that $x$ belongs to the category
represented by $f$). Given our limited capacity to
learn characters, we could take the environmental probability measure
$Q$ to have support on at most 10,000 distinct elements of $\F$ (so
$\F$ must contain more than just the 26 English letters).  Then if $x$
and $x'$ are two images of the same character, $f(x) \approx f(x')$
for all $f$ in the support of $Q$ and so by \eqref{rho}, $\rho(x,x')
\approx 0$, as required. If $x$ and $x'$ are images of different
characters, the classifiers $f_x$ and $f_{x'}$ corresponding to $x$
and $x'$ would have $f_x(x) \approx 1, f_x(x') \approx 0$ and
$f_{x'}(x) \approx 0, f_{x'}(x') \approx 1$. Classifiers for
characters that are subjectively similar to $x$ ($x'$) will behave in
a similar way to $f_x$ ($f_{x'}$), except that they will not assign
such a high value to positive examples as $f_x$ and $f_{x'}$ do. All
other classifiers $f$ will have $f(x) \approx f(x') \approx
0$. Substituting this into \eqref{rho} shows $\rho(x,x')$ will be
larger if $x$ and $x'$ are images of different characters than if they
are images of the same characters.

Note that $\rho$ depends only upon the environment $(\F,Q)$ and not
upon $X$ or its probability measure $P$.  Thus problems with the same
$X$ but different environments (for example character classification
and face recognition---different environments for the space of images)
will generate different canonical distortion measures.  In the next
section we will show that $\rho$ is the optimal distortion measure to
use if piecewise constant approximation of the functions in the
environment is the aim. Thus the fact that different environments
generate different $\rho$'s shows that the optimal similarity measure
between data points is highly dependent on what one is planning to do
with the data.

The definition of $\rho$ does not restrict us to considering only
classifier environments, any kind of functions will do. For example,
the environment might consist of all bounded linear functions on some
compact subset of $X=\R^n$, all threshold functions or all polynomials
up to a certain degree.  Noisy environments can also be modelled
within this framework by making the functions $f\in\F$ {\em
distribution} valued. For example, if the problem is to learn noisy
concepts then these can be represented as a distribution $P$ on $X$
combined with a conditional distribution on
$\{0,1\}$. Different concepts would have different conditional
distributions, which we can represent as mappings $f\colon X\to Y$
where now $Y$ is the set of all {\em distributions} on $\{0,1\}$.  The
distance measure $\sigma$ is now a distance measure on probability
distributions, such as Kullback-Liebler divergence or Hellinger distance.
The CDM between two inputs $x$ and $x'$ would
then equal the average distance between the {\em distributions} $f(x)$
and $f(x')$.

\section{Examples calculatable in closed form}
\label{examples}
\subsection{A linear environment}
\label{linear}
Suppose that $X=\R^n$ and $\F$ consists of all linear maps from $X$
into $\R$. $\F$ is the vector space dual of $X$ and so is itself
isomorphic to $\R^n$. With this in mind, take the measure $Q$ on $\F$
to be Lebesgue measure on $\R^n$, but restrict $Q$'s support to the
cube $[-\alpha,\alpha]^n$ for some $\alpha > 0$, and renormalise so
that $Q([-\alpha,\alpha]^n) = 1$.  Let $\sigma(y,y') = (y-y')^2$ for
all $y,y'\in\R$.  $\rho$ can then be reduced as follows:
\begin{eqnarray*}
\rho(x,x') &=& \int_\F\sigma(f(x),f(x'))\, dQ(f)\\ &=&
\frac1{(2\alpha)^n} \int\limits_{a\in [-\alpha,\alpha]^n} \(a\cdot x -
a\cdot x'\)^2\, d^na \\ &=& \frac1{(2\alpha)^n}
\int_{-\alpha}^\alpha\dots\int_{-\alpha}^\alpha \(\sum_{i=1}^n a_i x_i
- \sum_{i=1}^n a_i x'_i\)^2\, da_1\dots da_n \\* &=&
\frac{4\alpha^2}{3}\|x-x'\|^2.
\end{eqnarray*}
Thus a linear environment induces the squared Euclidean distance on
$X$. The reverse conclusion is also true, \ie if one assumes that
$\rho(x,x') = K\|x - x'\|^2$ for all $x,x'$ then $\F$ must be a linear
function class (almost everywhere). So based on the optimality
result of the next section, using the squared Euclidean distortion is
optimal if one wishes to approximate linear functions on the input
space, but is not optimal for any other environments.  As it is very
rare that one is interested in applying linear functions to images, or
speech signals (for example face classifiers are not linear maps on
image space, nor are word classifiers linear maps on speech signals),
the use of squared Euclidean distortion in these environments is not
the best thing to do.

Note that a uniform distribution over any {\em symmetric} region of
weight space will yield a CDM that is proportional to Euclidean
distance, while a general distribution will yield a CDM that is a
quadratic form $\rho(x,x') = \langle x|A|x'\rangle$ where $A$ is an
$n\times n$ matrix.

\subsection{A thresholded linear environment}
Take the same example as above but this time threshold the output of
each $f\in\F$ with the Heaviside step function, and take $Q$ to have
support only on the unit ball in $\R^n$, rather than the cube,
$[-\alpha,\alpha]^n$ (this is done to make the calculations simple).
After some algebra we find
$$
\rho(x,x') = \frac{\theta}{\pi},
$$
where $\theta$ is the angle between $x$ and $x'$. Thus in an
environment consisting of linear classifiers (\ie thresholded linear
functions) whose coefficients are distributed uniformly in the unit
ball, the natural distortion measure on the input space is the {\em
angle} between two input vectors.

\subsection{A quadratic environment}
\label{ex3}
Let $X=Y=[-1,1]$, $\sigma(y,y')=|y-y'|$ for all $y,y'\in Y$ and let
$\F=\{f\colon x\mapsto a x^2\}$ with $a$ uniformly distributed in the
range $[-1,1]$. With this environment,
\begin{eqnarray*}
\rho(x,y) &=& \int_{-1}^1 |a x^2 - a y^2|\, da \\* &=& |x-y| |x+y|.
\end{eqnarray*}
Note that $\rho(x,y)=0$ if $x=y$ and if $x=-y$, so that $x$ and $-x$
are zero distance apart under $\rho$.  This reflects the fact that
$f(x)=f(-x)$ for all $f\in\F$. Notice also that $\rho(x,y)$ is the
ordinary Euclidean distance between $x$ and $y$, {\em scaled} by
$|x+y|$. Thus two points with fixed Euclidean distance become further
and further apart under $\rho$ as they are moved away from the
origin. This reflects the fact that the quadratic functions in $\F$
have larger variation in their range around large values of $x$ than
they do around small values of $x$. This can also be seen by
calculating the $\ep$-ball around a point $x$ under $\rho$ (i.e the
set of points $x'\in X$ such that $\rho(x,x') \leq \ep$). To first
order in $\ep/{x}$ this is
$$
\[-x-\frac{\ep}{2 x}, -x + \frac{\ep}{2 x}\] \bigcup
\[x-\frac{\ep}{2 x}, x + \frac{\ep}{2 x}\].
$$
Note that the Euclidean diameter of the $\ep$-ball around $x$
decreases inversely linearly with $x$'s---Euclidean again---distance
from the origin.

\section{The optimality of the Canonical Distortion Measure}
\label{proof}
In this section it is shown that the CDM is the optimal distortion
measure to use if the goal is to find piecewise constant
approximations to the functions in the environment.

Piecewise constant approximations to $f\in\F$ are generated by
specifying a quantization $\xv=\{x_1,\dots,x_k\}$ ($x_i\in X$) of $X$ and
a partition $\Xv=\{X_1,\dots,X_k\}$ ($X_i\subseteq X, X_i\cap X_j =
\phi, X=\cup X_i$) of $X$ that is {\em faithful} to $\{x_1,\dots,x_k\}$
in the sense that $x_i\in X_i$ for $1\leq i\leq k$.  The piecewise
constant approximation $\fhat$ to any function $f$ is then defined by
$\fhat(x) = f(x_i)$ for all $x\in X_i$, $1\leq i\leq k$.

If information about the function $f$ is being transmitted from one
person to another using the quantization $\xv$ and the partition $\Xv$
then $\fhat$ is the function that will be constructed by the person on
the receiving end.

The most natural way to measure the deviation between $f$ and $\fhat$
in this context is with the pseudo-metric $d_P$,
$$
d_P(f,\fhat) := \int_X\sigma(f(x),\fhat(x))\,dP(x).
$$
$d_P(f,\fhat)$ is the expected difference between $f(x)$ and
$\fhat(x)$ on a sample $x$ drawn at random from $X$ according to
$P$. The {\em reconstruction error of $\F$} with respect to the pair
$\xv=\{x_1,\dots,x_k\}$ and $\Xv=\{X_1,\dots,X_k\}$ is defined to be
the expected deviation between $f$ and its approximation $\fhat$,
measured according to the distribution $Q$ on $\F$:
\begin{equation}
\label{define1}
E_\F(\xv,\Xv):= \int_\F d_P(f,\fhat)\, dQ(f).
\end{equation}
The quantization $\xv$ and partition $\Xv$ should be chosen so as to
minimize $E_\F(\xv,\Xv)$.

Given any quantization $\xv=\{x_1,\dots, x_k\}$ and distortion measure
$\rho$, define the partition $\Xv^\rho = \{X^\rho_1,\dots, X^\rho_k\}$
by 
$$
X^\rho_i := \{x\in X\colon \rho(x,x_i) \leq \rho(x,x_j),\,\text{for
all}\, j\neq i\}
$$ (break any ties by choosing the partition with the
smallest index).  Call this the partition of $X$ {\em induced} by
$\rho$ and $\xv$ (it is the Voronoi partition). Define
\begin{equation}
\label{define}
E_\F^\rho(\xv) := E_\F(\xv,\Xv^\rho).
\end{equation}

\begin{lem}
\label{twoerrs}
\begin{equation}
\label{equals}
E_\F^\rho(\xv) = E_\rho(\xv).
\end{equation}
\end{lem}
\sloppy{
{\bf Proof.}  Let $\xv=\{x_1,\dots,x_k\}$ be any quantization of $X$
and let $\Xv^\rho =\{X^\rho_1,\dots,X^\rho_k\}$ be the corresponding
partition induced by $\rho$. Denote the approximation of $f\in\F$ with
respect to this partition by $\fhat_\rho$. Then,
\begin{eqnarray*}
E_\F^\rho(\xv) &=& \int_\F d_P(f,\fhat_\rho)\,dQ(f) \\ &=&
     \int_\F\int_X \sigma(f(x),\fhat_\rho(x))\,dP(x)\,dQ(f) \\ &=&
     \int_X\int_\F \sigma(f(x),f(q_\rho(x)))\,dQ(f)\,dP(x) \\ &=&
     \int_X \rho(x,q_\rho(x))\, dP(x) \\* &=& E_\rho(\xv)
\end{eqnarray*}
$\qed$

\begin{thm}
\label{optimal}
The reconstruction error $E_\F$ of $\F$ is minimal with respect to a
quantization $\xv = \{x_1,\dots,x_k\}$ minimizing the reconstruction
error $E_\rho(\xv)$ of $X$, and the partition $\Xv^\rho$ induced by
the CDM $\rho$ and $\xv$.
\end{thm}
{\bf Proof.}  Let $\xv=\{x_1,\dots,x_k\}$ be any quantization of $X$
and let $\Xv^\rho =\{X^\rho_1,\dots,X^\rho_k\}$ be the corresponding
partition induced by $\rho$. Denote the approximation of $f\in\F$ with
respect to this partition by $\fhat_\rho$. Let $\Xv=\{X_1,\dots,X_k\}$
be any other partition of $X$ that is faithful to $\xv$ and let
$\fhat$ denote the approximation of $f$ with respect to this second
partition. Define $X_{ij} = X_i \cap X^\rho_j, 1\leq i \leq k, 1\leq
j\leq k$. Note that the $X_{ij}$'s are also a partition of $X$. The
reconstruction error of $\F$ with respect to the partition $\Xv$
satisfies:
\begin{eqnarray*}
E_\F(\xv,\Xv) &=& \int_\F d_P(f,\fhat)\, dQ(f) \\ 
&=& \int_\F\int_X
     \sigma(f(x),\fhat(x))\,dP(x)\,dQ(f) \\ &=&
     \sum_{i,j=1}^k\int_{X_{ij}}\int_\F\sigma(f(x),f(x_i))\,dQ(f)\,dP(x)
     \\ &=&
 \sum_{i,j=1}^k\int_{X_{ij}}\rho(x,x_i)\,dP(x) \\ &\geq&
     \sum_{i,j=1}^k\int_{X_{ij}}\rho(x,x_j)\,dP(x) \\ &=& \int_\F
     d_P(f,\fhat_\rho)\, dQ(f) \\* &=& E_\F^\rho(\xv).
\end{eqnarray*}
The theorem now follows from lemma \ref{twoerrs}.  $\Box$

Theorem \ref{optimal} states that as far as generating piecewise
constant approximations to the functions in the environment is
concerned, there is {\em no better} partition of the input space than
that induced by the CDM and its optimal quantization set.
}
\subsection{Quadratic environment revisited}
\label{qex}
\begin{figure}
\begin{center}
\leavevmode\epsfxsize=3in\epsfbox{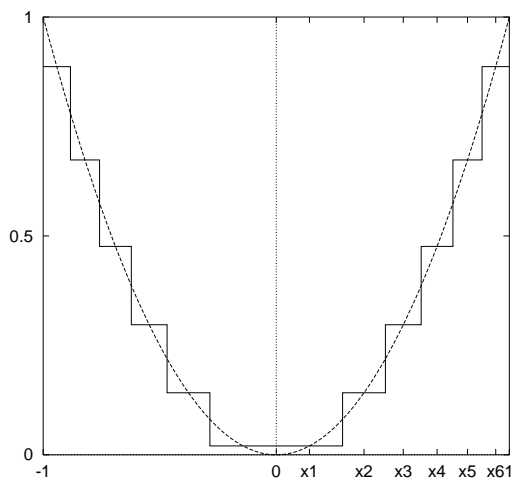}
\caption{Optimal quantization points and the corresponding
approximation to $f(x)=x^2$ for the quadratic environment of example
\ref{ex3}. The six quantization points are (to three significant
figures) $x_1=0.142$, $x_2=0.377$, $x_3=0.545$, $x_4=0.690$,
$x_5=0.821$, $x_6=0.942$.
\label{fhat}}
\end{center}
\end{figure}

For the quadratic environment of example \ref{ex3}, the optimal
quantization for $k=6$ is shown in figure \ref{fhat} along with $f$
and $\fhat_\rho$ for $f(x)=x^2$.  Note how the optimal quantization
reduces the deviation between $f$ and its approximation $\fhat_\rho$
by spacing the points closer together for larger values of $x$.

To calculate the optimal quantization in this case, first note that by
the symmetry of the environment, the quantization points $\{x_1,\dots,
x_6\}$ can all be assumed to be positive, and without loss of
generality suppose that $x_1\leq x_2 \leq\dots\leq x_6$. Direct
calculation of the reconstruction error 
$E_\rho(\xv)$ \eqref{recerr} shows that 
$$
x_i^2 = \frac14\(x_{i-1}^2 + x_{i+1}^2\) +
\frac1{4\sqrt{2}}\sqrt{x_{i-1}^4 + 6 x_{i-1}^2 x_{i+1}^2 + x_{i+1}^4}.
$$
A similar procedure can be used to show that $x_1$ and $x_6$ must
satisfy,
\begin{eqnarray*}
x_1 &=& \frac{x_2}{\sqrt{7}},\\* x_6 &=& \frac{4+\sqrt{2+7 x_5^2}}{7}.
\end{eqnarray*}
Optimal quantization points can be found by solving these equations
numerically. 

\section{Relationship between the CDM and other distance measures}
\label{relation}
\subsection{Transformation invariant distance}
The authors of \cite{Simard} introduced a technique for comparing
handwritten characters called {\em Transformation Distance}. They
observed that images of characters are invariant under transformations
such as rotation, dilation, shift, line thickening and so on. Denoting
the set of all such transformations by $G$, and assuming that $G$ can be
parameterised by $k$ real parameters, they noted that for each
character $x$, the set 
$$
\M(x) := \{gx\colon g\in G\}
$$
forms a $k$-dimensional manifold in the input space $X$. They defined
the distance bewteen two images $x$ and $x'$ to be:
\begin{equation}
\label{trans}
D(x,x') := \inf_{y\in \M(X),y'\in \M(x')} \|y - y'\|,
\end{equation}
that is, $D(x,x')$ is the smallest Euclidean distance between any
transformed image of $x$ and $x'$ (and is called the {\em
transformation distance between $x$ and $x'$}). In order to simplify
the computation, in \cite{Simard} $D(x,x')$ was approximated by
a linearised version. However we will concentrate on the exact expression  
\eqref{trans}.

Relating the transformation distance to the CDM, note that invariance
of characters under the action of $G$ is equivalent to assuming that
all character classifiers in the environment $\F$ are invariant under
the action of $G$. In other words, for all $f\in\F$, $f(gx) = f(x)$
for all $g\in G$. Then clearly, if $x'\in \M(x)$, $\rho(x,x') = D(x,x') =
0$, so the CDM gives the same answer as the transformation distance in
this case. However, if $x'\notin \M(x)$ then it is unlikely that
$D(x,x')$ and $\rho(x,x')$ will be the same, because $D(x,x')$
(somewhat arbitrarily) measures the Euclidean distance between the
manifolds $\M(x)$ and $\M(x')$, whereas $\rho(x,x')$ performs an
average over the functions in the environment to compute the distance.

The transformation distance and the CDM share the important property
that the distance between any pair of points in $\M(x)$ and $\M(x')$ 
respectively is always the same. This is trivial by the construction
of $D$; to see it for the CDM let $x,x'\in\M(x)$ and $y,y'\in M(y)$.
Then $\rho(x,x') = \rho(y,y') = 0$ and combining this with the 
triangle inequality gives:
\begin{eqnarray*}
\rho(x,y) &=& \rho(x,y) + \rho(x,x') + \rho(y,y') \\
  	&\geq& \rho(x',y) + \rho(y,y') \\*
	&\geq& \rho(x',y').
\end{eqnarray*}
Running the same argument with $x$ and $x'$ and $y$ and $y'$
interchanged
shows that $\rho(x,y) = \rho(x',y')$ always. 

\subsection{The CDM and Edelman's Chorus of Prototypes}
In \cite{Edelman1}, Edelman introduced a concept of representation he
called the {\em Chorus of Prototypes}. The idea is to train a set of
real-valued classifiers, $f_1,\dots,f_N$, for a domain of {\em
prototype} objects (so that $f_i(x)\in [0,1]$ is interpreted as the
probability that $x$ is an example of object $i$). All objects (not
just the prototypes) are then represented by the vector of activations
they induce at the output of the prototype classifiers. This vector of
activations is a {\em Low Dimensional Representation} (LDR) of the input
space. It is argued that the Euclidean distance between two
representation vectors corresponds to the distal similarity of the
objects.

Given that both the CDM and the Chorus of Prototypes represent
similarity by making use of an environment of classifiers, it is
natural to look for a connection between the two. There is a
connection if one assumes that all functions in the environment can be
implemented as linear maps composed with a fixed low dimensional
representation. So let $h = (h_1,\dots,h_k) \colon \R^d\to \R^k$ ($d
\gg k$) be the fixed LDR, and then each $f\in \F$ is of the form $\sum
w_i h_i$ where the $w_i$ depend on $f$. Suppose that $k$ is
minimal. Note that the $f$ take values outside the range $[0,1]$ and
so cannot be interpreted as probabilities. However we can still
interpret the output of $f$ as a ``degree of classification''---large
positive values being high confidence and large negative values low
confidence. In this case the environmental distribution $Q$ is a
distribution over the weights $w$. If $Q$ is uniform over a symmetric
region of weight space then by section \ref{linear}, the CDM between
two inputs $x$ and $x'$ will be proportional to the Euclidean distance
between their transformed representations $h(x)$ and $h(x')$.

Now choose any $f_1,\dots, f_k$ such that their respective weight vectors
$w^1,\dots,w^k$ are linearly independent (such a set of functions can
always be found because we assumed $k$ is minimal). $(f_1,\dots,f_k)$
will be the chorus of prototypes representation. Set 
\begin{equation}
W = 
\begin{array}{ccc}
w_1^1 & \dots & w_1^k \\
\vdots & \ddots & \vdots \\
w^1_k & \dots & w^k_k 
\end{array}
\end{equation}
and note that $W$ is nonsingular. For any input $x$, $(f_1(x),
\dots,f_k(x))$ (the representation of $x$ by its similarity to the
prototypes $i=1,\dots, k$) is equal to $W h(x)$.  Let $f = w\cdot h$
be any classifier in the environment. Set $w':= w W^{-1}$. Then $f =
w' W h = \sum w'_i f_i(x)$. Thus any classifier in the environment is
representable as a linear combination of prototype classifiers, as
required for the chorus of prototypes idea.

\subsection{Similarity measure of Thrun and Mitchell}
The authors of \cite{Thrun} defined an ``invariance function''
$\sigma\colon X\times X\to \{0,1\}$ for a finite environment $\F$ with
the property that if there exists $f\in\F$ such that $f(x) = 1$ then
$f'(x) = 0$ for all other $f'\in \F$:
$$
\sigma(x,x') := \left\{ \begin{array}{ll}
			1& \mbox{if $\exists f\in\F$ with $f(x)= f(x') = 1$} \\
			0& \mbox{if $\exists f\in\F$ with $f(x) \neq f(x')$} \\
			\mbox{undefined}& \mbox{otherwise}
			\end{array} \right.	
$$
If we assume the environmental distribution $Q$ is uniform on $\F$,
a quick calculation shows that for all $x,x'$ for which
$\sigma(x,x')$ is defined,
$$
\sigma(x,x') = 1 - \frac{|\F|}{2} \rho(x,x').
$$ 
Thrun and Mitchell also showed how $\sigma$ can be used to facilitate
learning of novel tasks within their {\em lifelong learning framework}.

\section{Learning the CDM with neural networks}
\label{learn}

For most environments encountered in practise (e.g speech recognition
or image recognition), $\rho$ will be unknown. In this section it is
shown how $\rho$ may be estimated or {\em learnt} using feed-forward
neural networks.  An experiment is presented in which the CDM $\rho$
is learnt for a toy ``robot arm'' environment. The learnt CDM is then used to
generate optimal Voronoi regions for the input
space, and these are compared with the Voronoi regions of the true CDM 
(which can be calculated exactly for this environment). 
Piecewise-constant approximations to the functions in the
environment are then learnt with respect to the Voronoi partition, and
the results are compared with direct learning using feedforward
nets. We conclude that learning piecewise-constant approximations
using the CDM gives far better generalisation performance than the
direct learning approach, at least on this toy problem. It should be
emphasised that this problem is designed primarily to illustrate the
ideas behind learning the CDM, and not as a practical test of the
theory on real-world problems. The latter awaits more sophisticated
experimentation. 

\subsection{Sampling the environment} 

To generate training sets for learning the CDM, 
both the distribution $Q$ over the environment $\F$ and the distribution 
$P$ over the input space $X$ must be sampled.
So let $\{f_1,\dots,f_M\}$ be $M$ \iid samples from $\F$ according to $Q$ 
and let $\{x_1,\dots,x_N\}$ be $N$ \iid samples from $X$ according to $P$.
For any pair $x_i,x_j$ an estimate of $\rho(x_i,x_j)$ is given by 
\begin{equation}
\label{rhat}
\rhat(x_i, x_j) = \frac1M\sum_{i=1}^M \sigma(f_i(x_i), f_i(x_j)).
\end{equation}
This generates $N^2$ training {\em triples},
$$\{(x_1,x_1,\rhat(x_1,x_1)), (x_1,x_2,\rhat(x_1,x_3)),\dots, (x_N, x_N,
\rhat(x_N, x_N))\},$$ 
which can be used as data to train a neural network. That is,
the neural network would have two sets of inputs---one set for $x_i$ and one
set for $x_j$---and a real-valued output $\rho^*(x_i,x_j)$ 
representing the network's estimate of $\rhat(x_i,x_j)$. The mean-squared 
error of the network on the training set is then 
\begin{equation}
\label{emp}
E = \frac1{N^2}\sum_{i=1}^N\sum_{j=1}^N 
	\[\rho^*(x_i,x_j) - \rhat(x_i,x_j)\]^2.
\end{equation}
$E$ is an estimate of the true distance between the network's 
$\rho^*$,  and the true CDM $\rho$, where this is defined by:
\begin{equation}
\label{true}
d(\rho,\rho^*) := \int_{X^2}\[\rho(x,x') - \rho^*(x,x')\]^2\,dP(x)\,dP(x').
\end{equation}

Note that the process of sampling from $\F$ to generate
$f_1,\dots,f_M$ is a form of multi-task learning (see \eg
\cite{colt95,caruana1,Thrun,pratt}) and that such sampling is a
necessary condition for the empirical estimate of the CDM, $\rho^*$,
to converge to the true CDM $\rho$.

\subsection{``Robot arm'' experiment}
\begin{figure}
\begin{center}
\leavevmode
\epsfxsize=3.0in\epsfbox{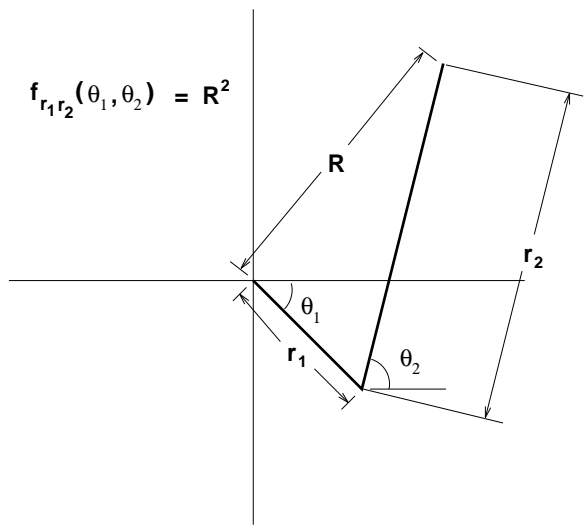}
\caption{\label{robot} The robot arm environment is generated by all two 
link robot arms with link lengths $0\leq r_1 \leq 1$ and $0\leq r_2 \leq 1$.
For each pair of angles $(\theta_1,\theta_2)$, 
$f_{r_1,r_2}(\theta_1,\theta_2)$ is the square of the distance of the end 
of the robot arm from the origin.}
\end{center}
\end{figure}

An artificial environment was created to test the effectiveness of
training neural networks to learn the CDM.  The environment was chosen
to consist of a set of two link ``robot arm'' problems. That is, each
function in the environment corresponded to a robot arm with two links
of length $r_1$ and $r_2$ (see Figure \ref{robot}. Note that the term 
``robot'' is used fairly loosely here: the example doesn't have much to do
with robotics). The function
corresponding to $r_1,r_2$ is the map $f_{r_1,r_2}
\colon [-\pi,\pi]^2 \to [0,(r_1+r_2)^2]$ 
that computes the square of the 
distance of the end of the arm from the origin, given the
angles of the links $\theta_1,\theta_2$.
Thus, $f_{r_1,r_2}(\theta_1,\theta_2) = r_1^2 + r_2^2 + 2r_1r_2 
\cos(\theta_1 - \theta_2)$. 
The link lengths $r_1$ and $r_2$ were chosen uniformly in the range $[0,1]$,
so that $\F = \{f_{r_1,r_2}\colon (r_1,r_2) \in [0,1]^2\}$. The goal
was to train a neural network to correctly implement the CDM 
$\rho((\theta_1,\theta_2),
(\theta'_1,\theta'_2))$. Note that in this case $\rho$ can be calculated 
in closed form:
\begin{equation}
\label{rhoex}
\rho((\theta_1,\theta_2),(\theta'_1,\theta'_2)) = 
\frac49\[\cos(\theta_1-\theta_2) - \cos(\theta'_1-\theta'_2)\]^2.
\end{equation}

The network architecture used was a single hidden layer neural network with 
$\tanh$ activation function on the hidden layer nodes and a linear 
output node. After some experimentation twenty hidden nodes was found to
be sufficient.
The network had four inputs, one each for $\theta_1,\theta_2,\theta'_1$ and 
$\theta'_2$. The knowledge that $\rho$ is symmetric was built into
the network by taking the output of the network for inputs $x_i$ and
$x_j$ to be 
\begin{equation}
\label{sym}
\frac{\rho^*(x_i,x_j) + \rho^*(x_j,x_i)}{2}
\end{equation}
rather than just the ``raw'' network output $\rho^*(x_i,x_j)$.
Note that 
\eqref{sym} is automatically symmetric in $x_i$ and $x_j$. With this choice
of the estimate of $\rho$, the error measure  \eqref{emp} becomes
\begin{equation}
\label{Ed}
E = \frac{2}{N(N+1)} \sum_{i=1}^N\sum_{j=i}^N 
\[\frac{\rho^*(x_i,x_j) + \rho^*(x_j,x_i)}2 - \rhat(x_i,x_j)\]^2.
\end{equation}
Back-propagation was used to compute the gradient of \eqref{Ed}. 

Training sets were generated by first sampling $M$ times from the environment 
to generate $f_1,\dots,f_M$, which in this case meant generating M pairs
$(r_1,r_2)$ uniformly at random in the square $[0,1]^2$. 
Then the input space
$[-\pi,\pi]^2$ was sampled uniformly at random $N$ times to generate 
$x_1,\dots,x_N$ and the empirical
estimate $\rhat(x_i,x_j)$ was constructed using \eqref{rhat} for each of the
$N(N+1)/2$ pairs
$(x_i,x_j)$ for $1\leq i\leq j\leq N$. 
A separate cross-validation set was also generated using the same values 
of $M$ and $N$ used for the training set.
The network was trained using conjugate gradient
descent---with the gradient computed from \eqref{Ed}---until 
the error on the cross-validation set failed to decrease for more
than five iterations in a row.
After the network had been trained, an initial quantization set
${q_1,\dots,q_m}$ of size $m\ll N$ was chosen uniformly at random from
$\{x_1,\dots,x_N\}$ and then the empirical Lloyd algorithm
\cite{lloyd} was used to optimize the positions of the quantization
points. The trained neural network---suitably symmetrised via
\eqref{sym}---was used as the distortion measure.

\begin{figure}
\begin{center}
\leavevmode
\epsfxsize=2.0in\epsfysize=2.0in\epsfbox{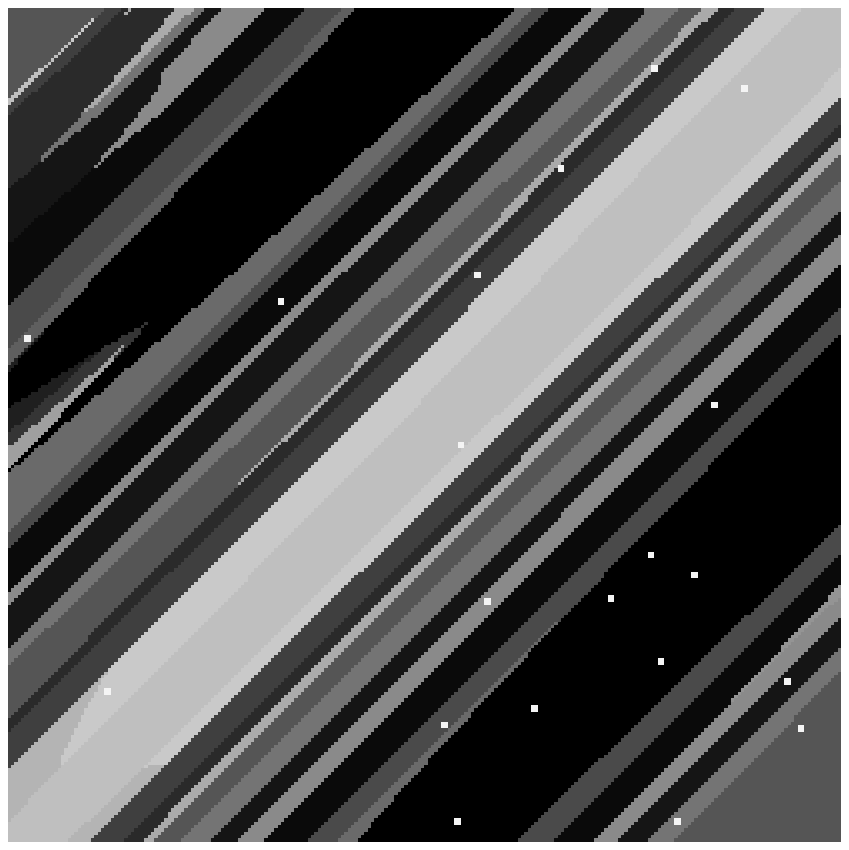}\hspace{5mm}
\epsfxsize=2.0in\epsfysize=2.0in\epsfbox{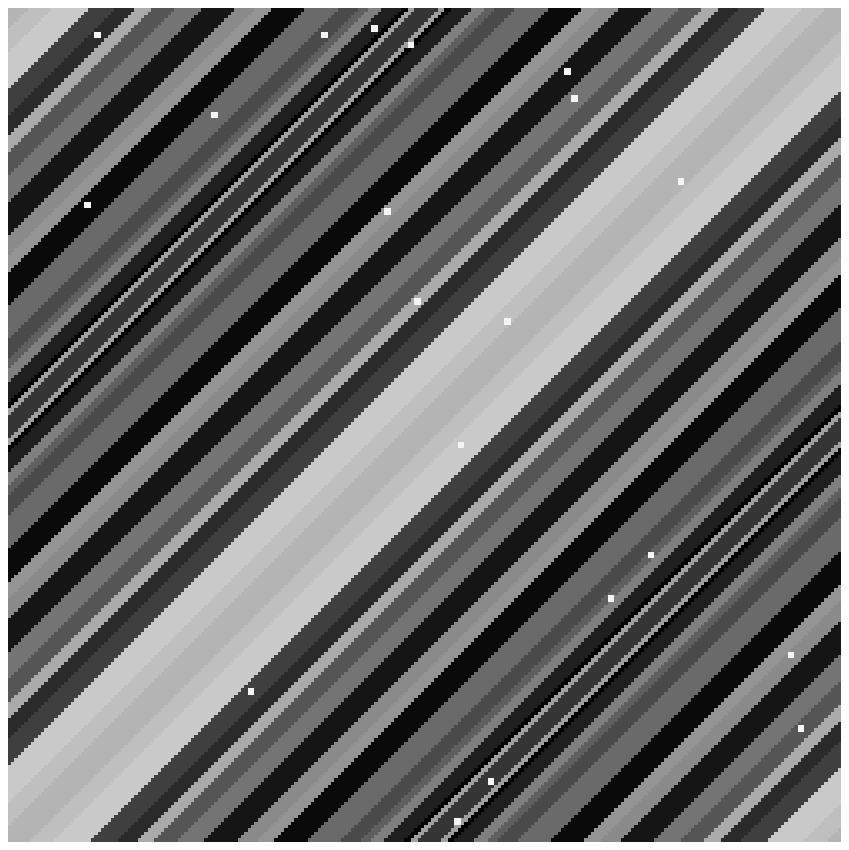}
\caption{\label{voronoi}
The picture on the left is the Voronoi diagram generated from $20$
quantization points by a neural network approximation to the CDM
($\rho^*$), trained on a sample size of $M=100$ functions and $N=100$
input points. The right-hand-side is the Voronoi diagram generated
using the true CDM.  All pixels with the same colour belong to the
same Voronoi region. The white dots are the quantization points found
by Lloyd's algorithm.}
\end{center}
\end{figure}

Several different experiments were performed with different values of $M,N$ and
$m$---the Voronoi regions for $(M,N,m) = (100,100,20)$ are plotted in
Figure \ref{voronoi}, along with the 
optimal quantization set and corresponding Voronoi 
regions for the true CDM \eqref{rhoex}. 
Note that the regions generated by the neural net
approximation are very similar to those generated by the true CDM. 
The reconstruction error of the input space was estimated from the
training set $\{x_1,\dots,x_N\}$ and also calculated (nearly) exactly
by quantizing the input space into a $250\times 250$ grid. These
quantities are shown in the first two columns of table \ref{rec} for
several different values of $M,N$ and $m$. Note the good agreement
between the estimate $\Ehat_{\rho^*}$ and the ``true'' value
$E_{\rho^*}$ in each case, and also that the reconstruction error is
very small in all cases, improving as the training set size ($M,N$)
for the CDM increases, and also as the number of quantization points
$m$ increases. 

\sloppy Recall that by lemma \ref{twoerrs},
$E_{\rho^*}(q_1,\dots,q_m)$ equals the reconstruction error of the
functions in the environment ($E^{\rho^*}_\F(q_1,\dots,q_m)$).  That is, if
\begin{enumerate}
\item a function $f$ is picked 
at random (\ie the link lengths $r_1$ and $r_2$ are picked 
at random),
\item  the values $f(q_i)$ for each $q_i$ in the optimal
quantization set are generated and stored, and
\item the value of $f$ at any novel input 
$x\in X$, chosen according to $P$, is estimated by $f(q_{\rho^*}(x))$,
\end{enumerate}
then $E_{\rho^*}(q_1,\dots,q_m)$ is the expected generalisation error:
$E_X\[f(x) - f(q_{\rho^*}(x))\]^2$.  Hence a small value of
$E_{\rho^*}(q_1,\dots,q_m)$ indicates that any function in the
environment is likely to be learnable to high accuracy by this
procedure\footnote{Note that this procedure is just $1$ Nearest
Neighbour, with $\rho^*$ as the distance metric and the quantization
points intelligently placed.}. To demonstrate this,
$E^{\rho^*}_\F(q_1,\dots,q_m)$ was estimated by generating 100 new
functions at random from the environment and learning them according to the
above procedure. Each function's generalisation error was estimated
using a fine grid on the input space, and then averaged across all
$100$ functions.  The average generalisation error is plotted in the
third column of table \ref{rec} for the various CDMs generated from
different training set sizes and the different numbers of quantization
points. Note the good agreement between the reconstruction error of
the environment ($\Ehat^{\rho^*}_\F$) and the reconstruction error of
the input space $E_{\rho^*}$.

\begin{figure}
\begin{center}
\leavevmode
\epsfxsize=3.0in\epsfbox{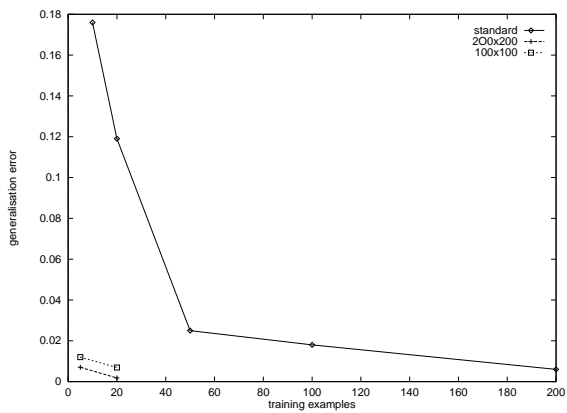}
\caption{\label{avgen}Average generalisation error as a function of 
training samples for the robot arm environment: standard approach vs. 
training using the estimated CDM. 
``Standard'' means learning each function from scratch with a one-hidden
layer, 10 hidden node neural net. ``$100\times 100$'' and 
``$200\times 200$'' refer to the number of functions and input examples
used to train the approximation to the CDM.}
\end{center}
\end{figure}

For comparison purposes the same 100 functions were learnt 
using a 10-hidden node neural network, without the assistance of the
CDM. The average generalisation error across all $100$ functions is 
displayed in Figure  \ref{avgen} along with the same quantity for the
piecewise-constant approximations. The piecewise-constant approach 
based on the estimated CDM is clearly superior to the normal approach 
approach in this case.

\begin{table}[h]
\caption{\label{rec}Empirical ($\Ehat_{\rho^*}$) and true
($E_{\rho^*}$) reconstruction error of the input space using Voronoi
regions and quantization points generated by the neural network
approximation ($\rho^*$) to the CDM. The final column is the estimated
reconstruction error of the functions in the environment.  $M$ is the
number of functions sampled from the environment and $N$ is the number
of points sampled from the input space to train $\rho^*$.  $m$ is the
number of quantization points.}
\begin{tabular*}{\textwidth}{@{\extracolsep{\fill}}ccccc} 
\hline
$(M,N)$ & $m$ & $\Ehat_{\rho^*}$ & $E_{\rho^*}$ & $\Ehat^{\rho^*}_\F$ \cr
\hline $(100,100)$ & $5$ & $8.3\times10^{-3}$&$1.0\times 10^{-2}$&$1.2\times 10^{-2}$ \cr
&$20$ & $6.5\times 10^{-4}$&$2.4\times 10^{-3}$&$6.9\times 10^{-3}$ \cr
\hline $(200,200)$ &$5$&$5.2\times 10^{-3}$&$5.0 \times 10^{-3}$&$7.9\times 10^{-3}$ \cr
&$20$&$6.0\times 10^{-4}$&$1.0\times 10^{-3}$&$1.8\times 10^{-3}$\cr
\hline
\end{tabular*}
\end{table}

\section{Conclusion}

It has been shown that the existence of an {\em environment} of
functions for a quantization process generates a {\em canonical
distortion measure} (CDM) $\rho$ on the input space.  It has been
proven that generating an optimal quantization set for the input space
using $\rho$ as the distortion measure automatically produces Voronoi
regions that are optimal for forming piecewise-constant approximations
to the functions in the environment.  The optimality theorem shows
that the CDM contains {\em all} the information necessary for learning
piecewise constant approximations to the functions in the
environment. Hence learning the CDM is a process of {\em learning to
learn}.

The CDM was calculated in closed form for several simple environments.
A surprising result is that the squared Euclidean distortion measure
is the CDM for a linear environment, and hence is optimal only if we
are interested in approximating linear functions.

Techniques for estimating the CDM and training a neural network to
implement it have been presented, and the results of several promising
experiments on a toy environment have been reported.  It remains to be
seen whether these techniques work for more complex domains such as
speech and character recognition.

One may be tempted to ask ``Why bother going to all the trouble of
learning the CDM first. Why not just learn the functions in the
environment directly?'' The answer to this question goes to the core
of what distinguishes learning to learn from ordinary learning. Of
course, if one is only interested in learning at most a handful of
functions from the same environment then learning the CDM is
overkill. Standard statistical learning techniques will be much more
efficient. However, if one is interested in in learning a large number
of functions from the environment, then learning the CDM will be a
big advantage, because by theorem \ref{optimal}, the CDM contains
{\em all} the information necessary for {\em optimal} learning of
piecewise constant approximations to the functions in the environment.

In ordinary learning we are interested primarily in aquiring the
information necessary to solve a particular learning problem, whereas
in learning to learn we want to acquire the information necessary to
solve a whole class of learning problems. As many real world problems
are members of large classes of similar problems (speech, face and
character recognition to name a few), tackling these problems from a
learning to learn perspective should be a fruitful approach.